\documentclass[10pt,twocolumn,letterpaper]{article}

\usepackage{cvpr}
\usepackage{times}
\usepackage{epsfig}
\usepackage{graphicx}
\usepackage{amsmath}
\usepackage{amssymb}
\usepackage{helvet}
\usepackage{courier}
\usepackage{url}
\usepackage[ruled]{algorithm2e}
\usepackage{caption}
\usepackage{subcaption}
\usepackage{color}
\usepackage{epstopdf}
\usepackage{tablefootnote}

\usepackage[pagebackref=true,breaklinks=true,letterpaper=true,colorlinks,bookmarks=false]{hyperref}

 \cvprfinalcopy 

\def\cvprPaperID{1425} 

\providecommand{\e}[1]{\ensuremath{\times 10^{#1}}}

\ifcvprfinal\fi
\begin{document}

\title{Hierarchical Recurrent Neural Encoder for Video Representation with Application to Captioning}

\author{Pingbo Pan$^\S$\hspace{1em}Zhongwen Xu$^\dag$\hspace{1em} Yi Yang$^\dag$\hspace{1em} Fei Wu$^\S$\hspace{1em} Yueting Zhuang$^\S$\\
	$^\S$Zhejiang University \hspace{1em}   $^\dag$University of Technology Sydney\\
	{\tt\small \{lighnt001,zhongwen.s.xu\}@gmail.com} \hspace{1mm} \tt \small yi.yang@uts.edu.au \hspace{1mm} \\{\tt\small \{wufei,yzhuang\}@cs.zju.edu.cn}
}

\maketitle

\begin{abstract}
Recently, deep learning approach, especially deep Convolutional Neural Networks (ConvNets), have achieved overwhelming accuracy with fast processing speed for image
classification. Incorporating temporal structure with deep ConvNets for video representation becomes a fundamental problem for video content
analysis. 
In this paper, we propose 
 a new approach, namely Hierarchical Recurrent Neural Encoder (HRNE), to exploit temporal information of videos. Compared to recent video representation inference approaches, this paper makes the following three contributions.  First, our HRNE is able to efficiently exploit video temporal structure in a longer range by reducing the length of input information flow, and compositing multiple consecutive inputs at a higher level. Second, computation operations are significantly lessened while attaining more non-linearity. Third, HRNE is able to uncover temporal transitions between frame chunks with different granularities, \ie it can model the temporal transitions between frames as well as the transitions between segments. We apply the new method to video captioning where temporal information plays a crucial role. Experiments demonstrate that our method outperforms the state-of-the-art on video captioning benchmarks. Notably, even using a single network with only  RGB stream as input, HRNE beats all the recent systems which combine multiple inputs, such as RGB ConvNet plus 3D ConvNet.
\end{abstract}
\section{Introduction}
Incorporating temporal information into video representation has long been a fundamental problem in computer vision. 
Earlier works such as Dense Trajectories~\cite{wang2011action} and improved Dense Trajectories (iDT)~\cite{wang2013action} typically utilize optical flow to extract temporal information and hand-crafted features to model appearances and motions.  
With the recent success of deep Convolutional Neural Networks (ConvNets) both in efficiency and efficacy, we have witnessed a new trend in leveraging ConvNets to infer 
video representation. Xu~\etal~\cite{xu2015discriminative} propose to utilize VLAD~\cite{jegou2010aggregating} to aggregate frame level ConvNet for video representation, which is unable to capture temporal structure. 
Simonyan and Zisserman~\cite{simonyan2014two} combine stacked optical flow frames and RGB streams to train ConvNets for video classification, which achieves comparable performance to iDT in action recognition. A limitation of two-stream ConvNets~\cite{simonyan2014two} and iDT~\cite{wang2013action} is that both algorithms require optical flow as input, which is expensive to extract (it takes usually 0.06 seconds to extract optical flow between a pair of frames~\cite{simonyan2014two}), but is  only able to capture temporal information in video clips of short duration.

To avoid extracting optical flow, 3D ConvNets are proposed in~\cite{DBLP:journals/corr/TranBFTP14} to generate a video representation, with emphasis on efficiency improvement. 
This approach, however, can only cope with short video clips of 16 frames or so~\cite{DBLP:journals/corr/TranBFTP14}.
Very recently, Long Short-Term Memory (LSTM)~\cite{hochreiter1997long} has been applied to video analysis~\cite{ng2015beyond}, inspired by the general recurrent encoder-decoder framework~\cite{sutskever2014sequence}. A plausible feature of LSTM is that LSTM is capable of modeling data sequences. 
However, as this paper tries to cope with, there are still a few challenges remain unaddressed.

First,  a large number of long-range dependencies are usually difficult to capture. Even though LSTM can deal with long video clips in principal, it has been reported that the favorable length of video clips to be feed into LSTM falls in the range of 30 to 80 frames~\cite{ng2015beyond,venugopalan2015sequence}. 
In order to model longer video clips while attaining similar good performance as in~\cite{ng2015beyond,venugopalan2015sequence}, we propose to divide a long video clip into a few short frame chunks, feed the chunks into LSTM, and  composite the LSTM outputs of the frame chunks into one vector, which can then be fed into another LSTM at a higher level to uncover the temporal information among the composited vectors over a longer duration. Such a hierarchical structure significantly reduces the length of input information flow but is still capable of exploiting temporal information over longer time afterwards at a higher level.

Second, additional non-linearity has been demonstrated helpful for improving model training for visual tasks such as image and video classification~\cite{Simonyan14c,sutskever2014sequence,ng2015beyond}.  
A straightforward way of adding non-linearity into LSTM is stacking~\cite{sutskever2014sequence,ng2015beyond}. Despite of the improved performance, a major disadvantage of stacking is that it introduces a long path from the input to the output video vector representation, thereby resulting in heavier computational cost.  As we will discuss in details later, Hierarchical Recurrent Neural Encoder (HRNE) proposed in this paper dramatically shortens the path with the capability of adding non-linearity, providing a better trade-off between efficiency and effectiveness. 

Third, video temporal structures are intrinsically layered. Suppose a video of birthday party consists of three actions, \ie, blowing candles, cutting cake, and eating cake. As the three actions usually take place sequentially, \ie, there are strong temporal dependencies among them, we need to appropriately model the temporal structure among the 
three actions. In the meantime, the temporal structure within each action should also be exploited. To this end, we need to model video temporal structure with multiple granularities.  Unfortunately, straightforward implementation of LSTM can not achieve this goal.


The proposed HRNE framework models video temporal information using a hierarchical recurrent encoder and can effectively deal with the three aforementioned challenges.
While HRNE is a generic video representation, we apply it to video captioning to test the performance, because temporal information plays a key role in video captioning. Two widely-used video captioning datasets, the Microsoft Research Video Description Corpus (MSVD)~\cite{chen2011collecting} and the Montreal Video Annotation Dataset (M-VAD)~\cite{torabi2015using}, are used in our experiments, which demonstrate the effectiveness of the proposed method.  





\section{Related Works}
\label{sec:relatedwork}

Dense Trajectories~\cite{wang2011action} and its improved version: improved Dense Trajectories~\cite{wang2013action} have dominated the filed of action recognition and general video classification tasks such as complex event detection. Dense Trajectories applies dense sampling to get the interest points along the video and then tracks the points in a short time period. Local descriptors such as HOG, HOF and MBH are extracted along the tracklets. Bag-of-Words (BoWs)~\cite{sivic2003video} and Fisher vector encoding~\cite{sanchez2013image} are then applied to accumulate the local descriptors and generate the video representation. 

Besides the hand-crafted visual features like Dense Trajectories, researchers have started exploring the  Convolutional Neural Networks (ConvNets) on video representation recently. Karpathy~\etal~\cite{karpathy2014large} first introduce ConvNets which are similar with Krizhevsky~\etal~\cite{krizhevsky2012imagenet} into video classification, and different fusion strategies are explored to combine information over the temporal domain in this work. In order to better capture temporal information in action recognition, Simonyan and Zisserman~\cite{simonyan2014two} propose to utilize stacked optical flow frames as inputs to train the ConvNets, which, together with the RGB stream, achieves comparable performance as the state-of-the-art hand-crafted features~\cite{wang2013action} on action recognition. Tran~\etal~\cite{DBLP:journals/corr/TranBFTP14} utilize 3D ConvNets to learn temporal information without optical flows, which is inspired by Ji~\etal~\cite{ji20133d} and Simonyan and Zisserman~\cite{Simonyan14c}. Xu~\etal~\cite{xu2015discriminative} propose to utilize VLAD~\cite{jegou2010aggregating} aggregation on frame-level ConvNet features and it directly adapts ImageNet pretrained image classification model to video representation.

All these works mentioned above utilize either average pooling or encoding methods such as Fisher vector and VLAD over  time to generate a global video feature from a set of local features. However, time dependency information is lost since average pooling and encoding methods always ignore the order of the input sequences, \ie, taking the local features as a~\emph{set} rather than a~\emph{sequence}. To tackle this problem, Ng~\etal~\cite{ng2015beyond} introduce Long Short-Term Memory (LSTM) to model the temporal order, inspired by the general sequence to sequence learning neural model proposed by Sutskever~\etal~\cite{sutskever2014sequence}. Stacked LSTM is applied in~\cite{ng2015beyond}, where each layer of the LSTM retains the same time scale. Different from the standard approach which stackes LSTM layers into multilayered one and simply aims to introduce more non-linearity into the neural model, the \emph{Hierarchical Recurrent Neural Encoder} proposed in this work aims to abstract the visual information at different time scales, and learns the visual features with multiple granularities.

In the application of video captioning, Donahue~\etal~\cite{donahue2014long} introduce the LSTM into this task by feeding the Conditional Random Field (CRF) outputs of objects, subjects, and verbs into the LSTM to generate video description. The same as~\cite{donahue2014long}, other works such as~\cite{venugopalan2014translating, yao2015describing, pan2015jointly}   utilize the LSTM essentially as a recurrent neural network \emph{language model} to generate  video descriptions, which conditions on either the average pooled frame-level features or the context vector linearly blended by the attention mechanism~\cite{bahdanau2014neural}. In contrast to these works, we study better video content understanding 
 from the visual feature aspects instead of language modeling ones. Based on   stacked LSTM, Venugopalan~\etal~\cite{venugopalan2015sequence} is the only attempt to utilize LSTMs as both visual encoder and language decoder in the video captioning task, which is inspired by the general neural encoder-decoder framework~\cite{sutskever2014sequence,cho2015learning} as well.

In the area of query suggestion, Sordoni~\etal\cite{Sordoni2015hierarchical} propose a  hierarchical recurrent neural network for context-aware query suggestion in a search engine. In this model, the text query in a session is firstly abstracted by one RNN layer into the query-level state, another RNN layer is  used to learn session-level dependency and then, the session-level hidden states is utilized to make suggestions for users. 

 
Contemporary to this work, Yu~\etal~\cite{yu2015video} introduce a hierarchical RNN \emph{decoder}, specifically Gated Recurrent Unit (GRU)~\cite{cho2015learning}, into the video captioning system. A sentence generator consisting of a GRU layer conditions on  visual feature, and then a paragraph generator accepts sentence vector  and the context to generate paragraph level description, which essentially learns the time dependencies between sentences, and works on the language processing aspects. In contrast, this work is focusing on learning good \emph{visual feature}, \ie, the encoder part, but not the \emph{language processing}, \ie, the  decoder part.
\section{The Proposed Approach}
\label{sec:model}
We propose a Hierarchical Recurrent Neural Encoder (HRNE) model for video processing tasks. Assume we have $n$ frames in the video, based on the HRNE model, we develop a general video encoder which takes the frame-level visual features from a video sequence $(\mathbf{x}_1, \mathbf{x}_2, ..., \mathbf{x}_n)$ as input and outputs a single vector $\mathbf{v}$ as the  representation for the whole video. For the video captioning task specifically, we keep the single layer LSTM decoder as a recurrent neural network language model~\cite{sutskever2014sequence}, which conditions on the video feature vector $\mathbf{v}$, similar to previous works~\cite{venugopalan2015sequence,yao2015describing}. By keeping the similar LSTM decoder, we can make a fair comparison with other neural network video captioning systems and demonstrate the power of our hierarchical recurrent neural encoder for the visual information extraction.

\subsection{The Recurrent Neural Network}
\label{sec:rnn}
The recurrent neural network is a natural extension of feedforward neural networks on modeling sequence. Given an input sequence $(\mathbf{x}_1, \mathbf{x}_2 ..., \mathbf{x}_n)$, a standard RNN computes the output sequence $(\mathbf{z}_1, \mathbf{z}_2 ..., \mathbf{z}_n)$ by iterating the following equations:
\begin{gather}
 \mathbf{h}_t = \tanh(W_{hx}\mathbf{x}_t + W_{hh} \mathbf{h}_{t-1}) \\
 \mathbf{z}_t = W_{zh} \mathbf{h}_t
 \end{gather}

The RNN can map the inputs to the outputs whenever the alignment between inputs and outputs is provided. The standard RNN would work principally, but it is really difficult to train the standard RNN due to the vanishing gradient problem~\cite{bengio1994learning}. The Long Short-Term Memory (LSTM) is known to learn patterns with wider range temporal dependencies. We now introduce the LSTM model.

\begin{figure}[t!]
\includegraphics[width=0.5\textwidth]{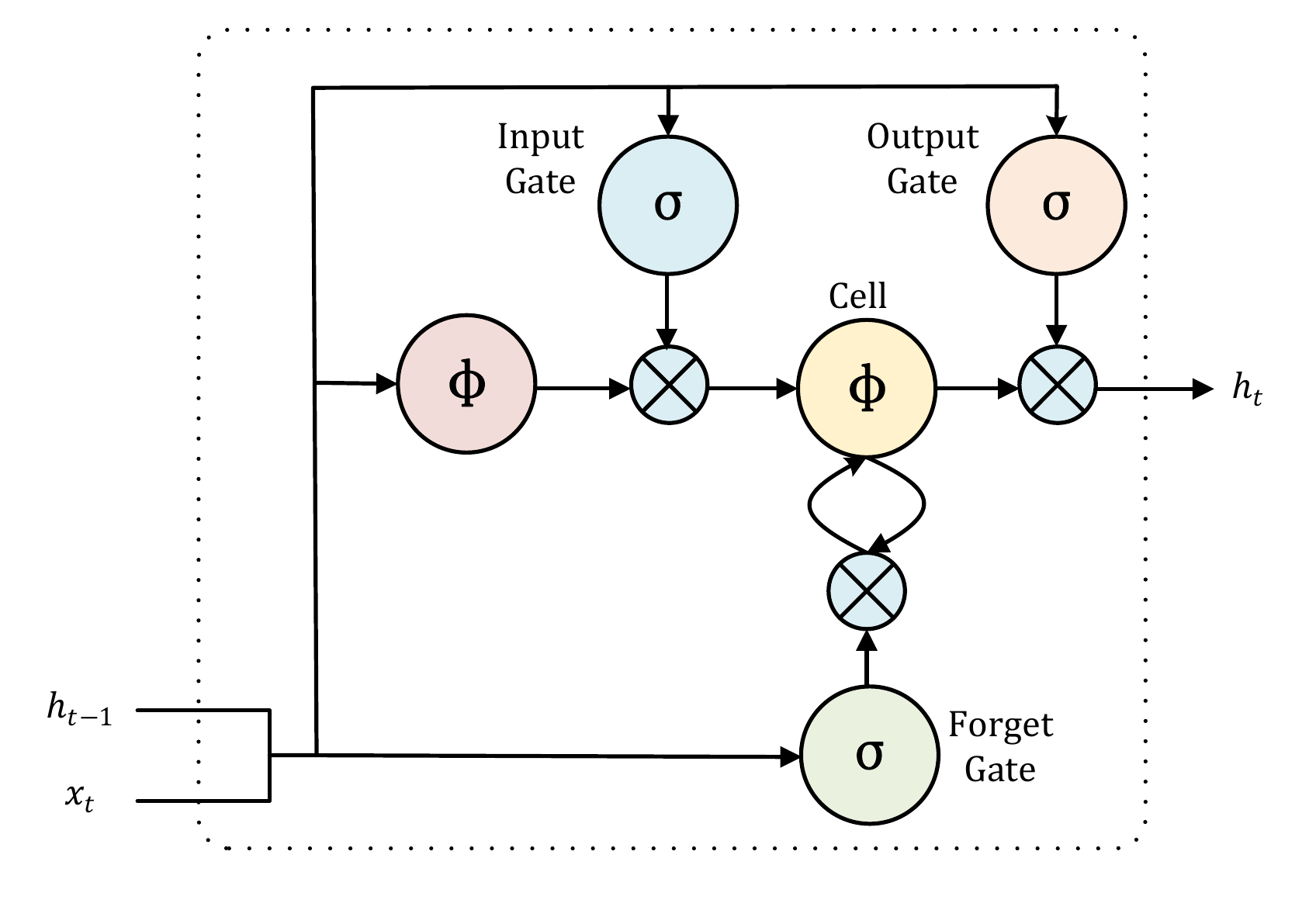}
\caption{An illustration of the LSTM unit, replicated from~\cite{donahue2014long}.} 
	
\label{fig:lstmunit}
\end{figure}

The core of the LSTM model is a memory cell $\mathbf{c}_t$ which records the history of the inputs observed up to that time step. $\mathbf{c}_t$ is a summation of the previous memory cell $\mathbf{c}_{t-1}$ modulated by a sigmoid gate $\mathbf{f}_t$, and $\mathbf{g}_t$, a function of previous hidden state and the current input modulated by another sigmoid gate $\mathbf{i}_t$. The sigmoid gates can be thought as knobs that LSTM learns to selectively forget its memory or accept current input. The cell has three gates. The input $\mathbf{i}_t$ gate controls whether the LSTM will consider current input $\mathbf{x}_t$. The forget gate $\mathbf{f}_t$ is used to control whether LSTM will forget the previous memory $\mathbf{c}_{t-1}$. The output gate $\mathbf{o}_t$ controls how much information will be transferred from memory $\mathbf{c}_t$ to hidden state $\mathbf{h}_t$. There are several widely used LSTM variants and we use the LSTM unit described in \cite{DBLP:journals/corr/ZarembaS14} (see also Figure~\ref{fig:lstmunit})  in our model, which iterates as follows:

\begin{figure*}[t!]
	\begin{subfigure}[h] {0.5\textwidth}
		\begin{center}
			\includegraphics[height = 0.5\textwidth]{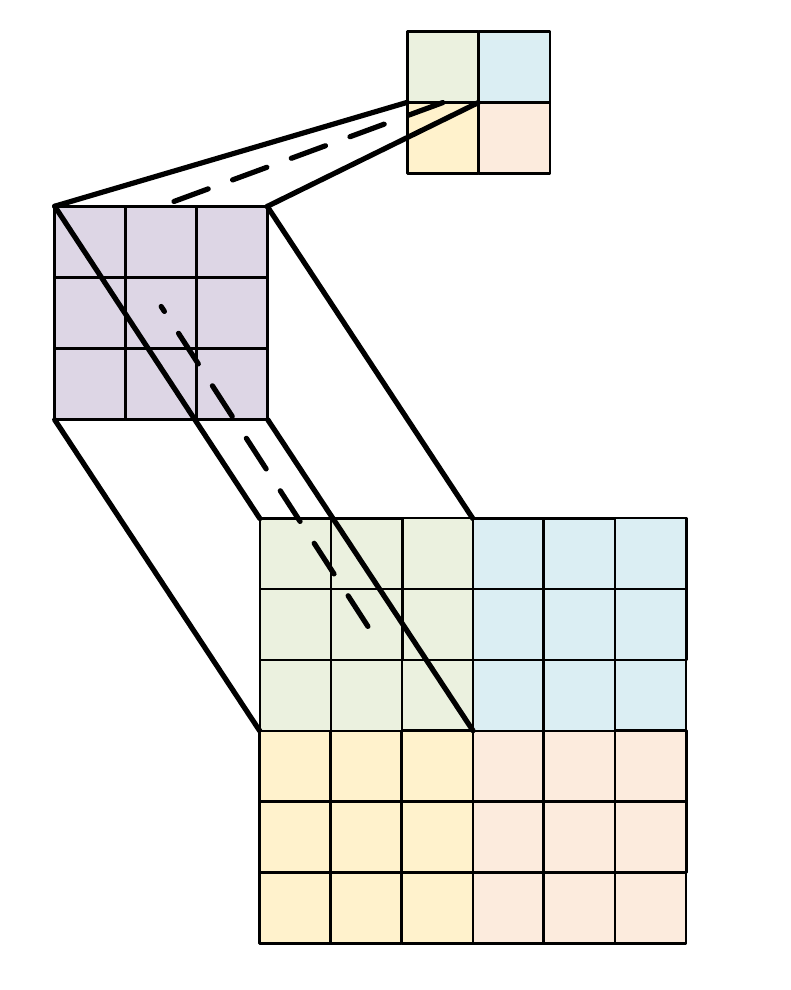}
			\caption{Spatial convolutional operations of ConvNet}
			\label{fig:cnnfilter}
		\end{center}
	\end{subfigure}\hspace{-0.6 cm}
	\begin{subfigure}[h] {0.5\textwidth}
		\begin{center}
			\includegraphics[height = 0.5\textwidth]{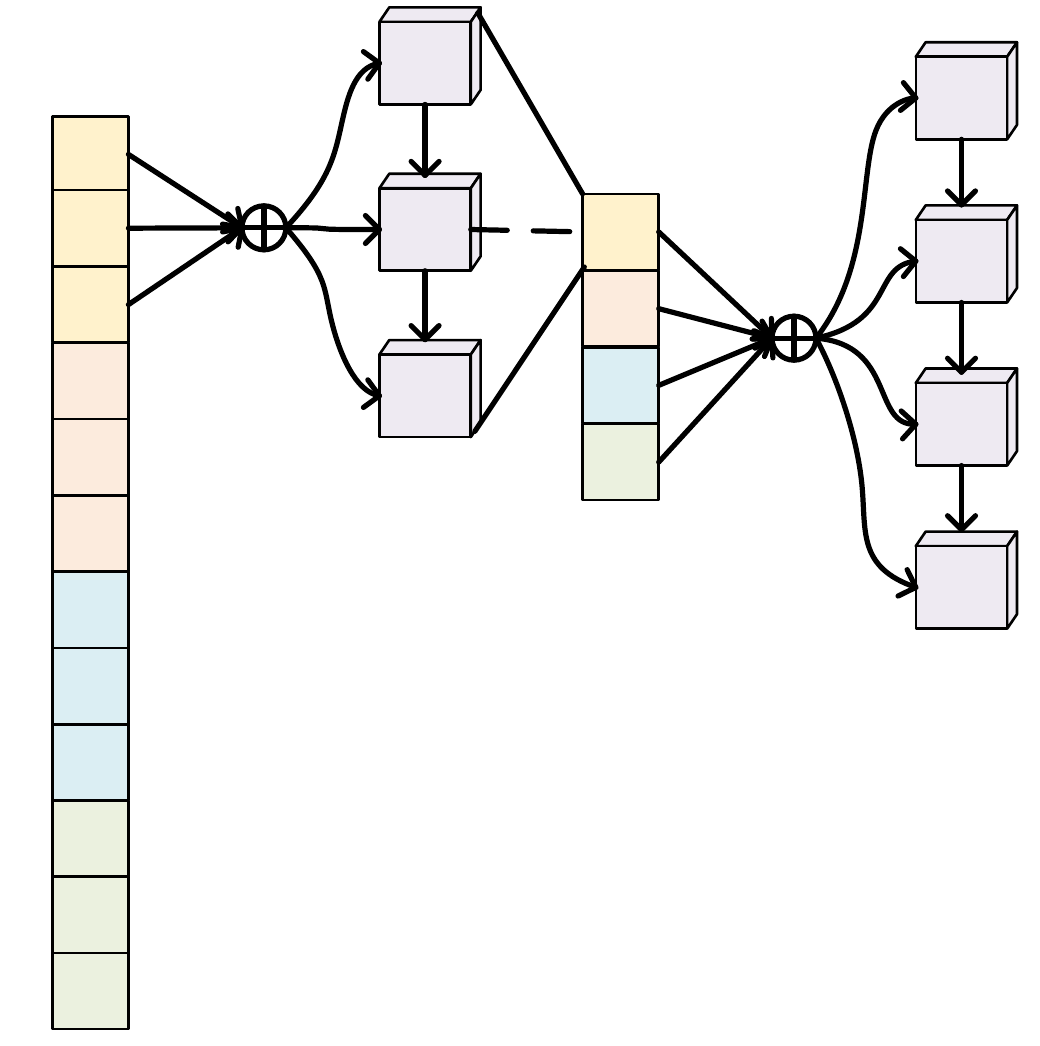}
			\caption{Temporal operations of Hierarchical Recurrent Neural Encoder}
			\label{fig:rnnfilter}
		\end{center}
	\end{subfigure}
	\caption{Analogical illustration of temporal operations of HRNE to spatial convolutional operations of ConvNet. In ConvNet a learnable filter is applied to each location to generate a filtered image which is further forwarded to the next layer. In HRNE, a learnable filter (\ie LSTM) along with attention mechanism is applied to each temporal time step to generate a sequence of video chunk vectors, which are further forwarded to the next layer. }
	\label{fig:filter}
\end{figure*}

\begin{gather}
\mathbf{i}_t = \sigma (W_{ix} \mathbf{x}_t + W_{ih} \mathbf{h}_{t-1} + \mathbf{b}_i)   \\
\mathbf{f}_t = \sigma (W_{fx} \mathbf{x}_t + W_{fh} \mathbf{h}_{t-1} + \mathbf{b}_f)   \\
\mathbf{o}_t = \sigma (W_{ox} \mathbf{x}_t + W_{oh} \mathbf{h}_{t-1} + \mathbf{b}_o)  \\
\mathbf{g}_t = \phi (W_{gx} \mathbf{x}_t + W_{gh} \mathbf{h}_{t-1} + \mathbf{b}_g)    \\
\mathbf{c}_t = \mathbf{f}_t \odot \mathbf{c}_{t-1} + \mathbf{i}_t \odot \mathbf{g}_t    \\
\mathbf{h}_t = \mathbf{o}_t \odot \phi (\mathbf{c}_t),
\end{gather}
where $\sigma$ is the sigmoid function, $\phi$ is the hyperbolic tangent function~$\tanh$, $\odot$ donates element-wise product, $W_{*x}$ is the transform from the input to LSTM states, $W_{*h}$ is the recurrent transformation matrix between the hidden states and $\mathbf{b}_i$ is the biases vector.

\subsection{Hierarchical Recurrent Neural Encoder}
It has been reported that adding more non-linearity is helpful for vision tasks \cite{Simonyan14c}. The performance of LSTM can be improved if 
additional non-linearity is added. A straightforward way is stacking multiple layers, which, however, will  increase computation operations.  
Inspired by the ConvNet operations in spatial domain, we propose a Hierarchical Recurrent Neural Encoder (HRNE) model. As shown in Figure \ref{fig:filter}, in a ConvNet model, a filter is used to explore the spatial visual information of an image by performing convolution calculation between image patch matrix $I$ and a learnable filter matrix $H$:
\begin{equation}
y = \sum_{i=1}^h\sum_{j=1}^w I_{i,j} H_{i,j},
\end{equation}
where $w$ denotes the number of columns of the filter matrix, $h$ denotes the number of rows, $H_{i,j}$ denotes the matrix item located in the $i$-th row and $j$-th column and $y$ is the convolution calculation result. The filter is applied over the whole image to generate the filtered image, which is further forwarded into the next layer. 
Similarly, in  temporal domain, we  introduce an additional layer, instead of stacking, by which only short LSTM chains need to be dealt with. Like the filters in ConvNet's convolutional layer are well suited for exploring local spatial structure, using temporal filter to explore the local temporal structure is presumed to be beneficial 
since videos always consist of several incoherence clips.

The main difficulty of introducing additional layers into temporal modeling is finding a proper temporal filter. In spatial domain,  the output of filter is independent from spatial location, and a matrix can be used as a filter. Differently, in temporal domain, there is certain temporal dependencies between consecutive items. As a result, a matrix is not sufficient to be used as a temporal filter. Since RNN is well suited for temporal dependency modeling, we adopt short RNN chains as the temporal filters in our HRNE model. Specifically, we  use LSTM chains in this paper and take the mean of all  LSTM chain's hidden states as the filtering result.


We first divide an input sequence $(\mathbf{x}_1, \mathbf{x}_2, \ldots, \mathbf{x}_T)$ into several chunks $(\mathbf{x}_1, \mathbf{x}_2, ..., \mathbf{x}_n)$, $(\mathbf{x}_{1+s}, \mathbf{x}_{2+s}, \ldots, \mathbf{x}_{n+s})$, \ldots, $(\mathbf{x}_{T-n+1}, \mathbf{x}_{T-n+2},\ldots, \mathbf{x}_T)$, where $s$ is stride and it denotes the number of temporal units two adjacent chunks are apart. After inputting these subsequences into the LSTM filter, we will get a  sequence of feature vectors $\mathbf{h}_1, \mathbf{h}_2, .., \mathbf{h}_{\lceil T/n \rceil}$, where $\lceil x \rceil$ denotes the least integer among those integers which are larger than $x$. Each feature vector in $\mathbf{h}_1, \mathbf{h}_2, .., \mathbf{h}_{\lceil T/n \rceil}$ gives a proper abstract of its corresponding clip. Since what we actually need is the feature vector of the whole video, we must figure out a method to summarize all these feature vectors. We propose to use another LSTM layer to deal with this task. We combine these two LSTM layers and build our HRNE model. The first LSTM layer serves as a filter and it is used to explore local temporal structure within subsequences. The second LSTM learns the temporal dependencies among subsequences. We note that more complex HRNE model could be adding more layers to build multiple time-scale abstraction of the visual information.

\begin{figure*}
\begin{subfigure}[h] {0.51\textwidth}
\begin{center}
\includegraphics[height = 0.5\textwidth]{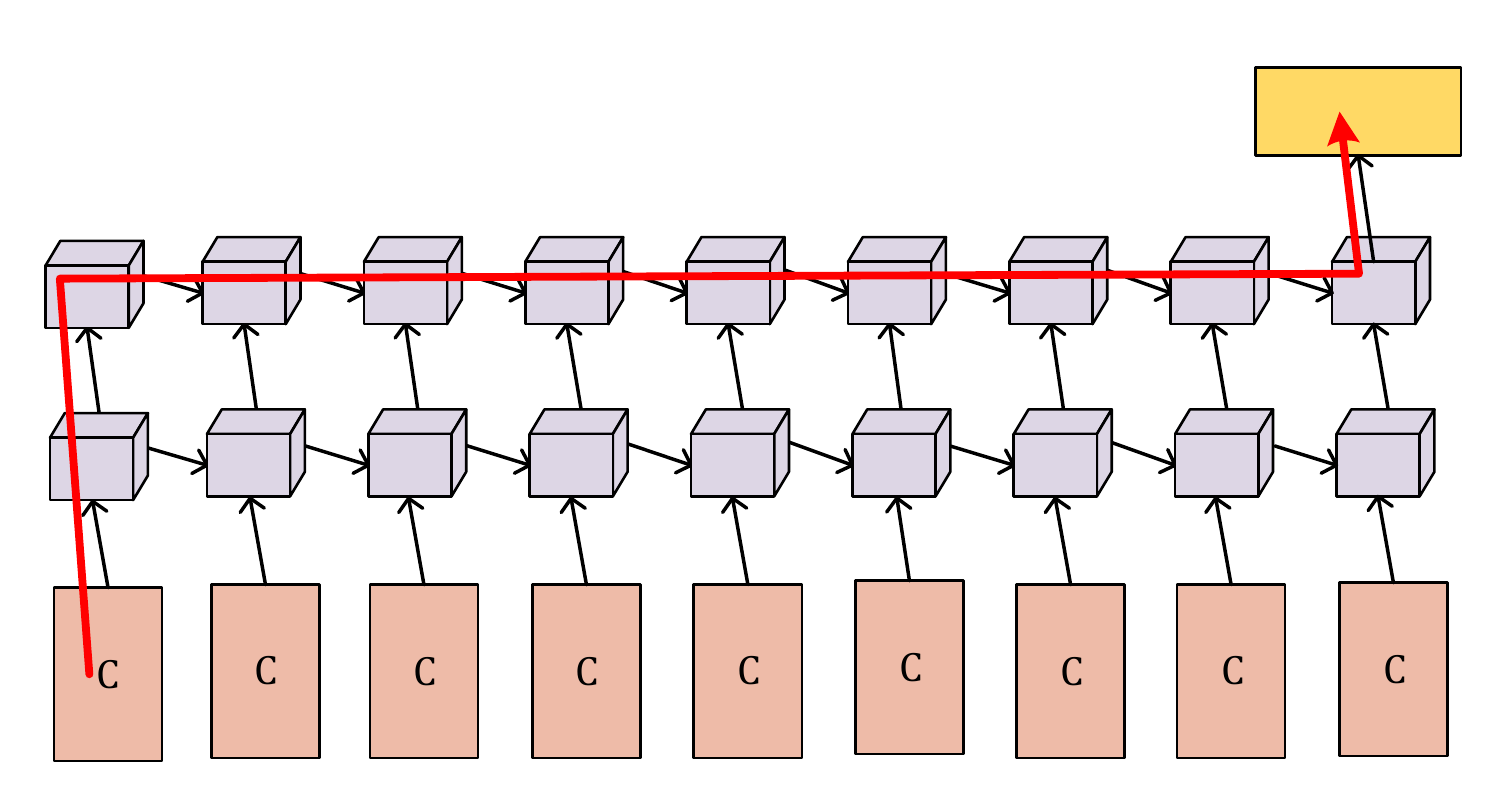}
\caption{Stacked LSTM video encoder}
\label{fig:deeprnn}
\end{center}
\end{subfigure}
\begin{subfigure}[h] {0.5\textwidth}
\begin{center}
\includegraphics[height = 0.5\textwidth]{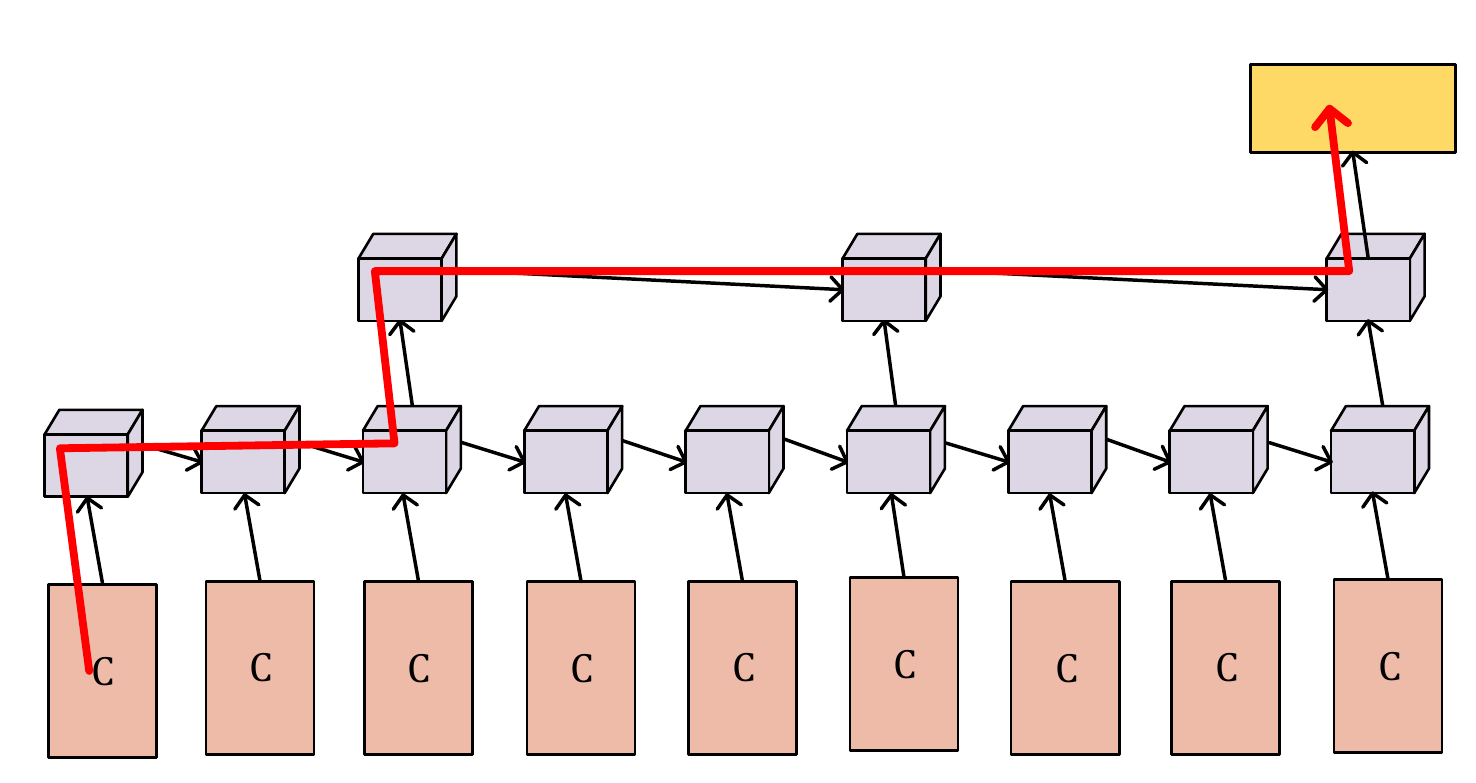}
\caption{Hierarchical Recurrent Neural Encoder}
\label{fig:convrnn}
\end{center}
\end{subfigure}
\caption{A comparison between stacked LSTM and the proposed Hierarchical Recurrent Neural Encoder. This figure takes a two layer hierarchy as an example to showcase. The red line in each subfigure shows one of the paths from the visual appearance input at $t=1$ to the output video vector representation. There are 10 time steps in stacked LSTM and only 6 time steps in our model.  }
\label{fig:rnn}
\end{figure*}

A large number of long-range dependencies are usually difficult to capture. Even though LSTM can deal with long video clips in principal, we compare HRNE with stacked multilayered LSTM in Figure \ref{fig:rnn}. The red line in the Figure~\ref{fig:rnn} shows how the input at $t=1$ flows though the model to the final output. We are used to set the stride to be the same as the LSTM filter length. For an input sequence of length $T$ and a LSTM filter of length $n$, the red line in HRNE model goes through $n + \lceil T/n \rceil$ LSTM units, which means the input at $t=1$ will only flow through $n + \lceil T/n \rceil$ steps to the output rather than $T+1$ steps if stacked RNN is used. If $T = 1,000$ and we set $n$ to be 30, then HRNE will only go through 64 steps rather than 1,001 steps. Fewer steps an input will go through before it reaches the output means that it's easier to backtrack, so our HRNE is easier for stochastic gradient methods via Back-propagation Through Time (BPTT) to train.

Since the recently proposed soft attention mechanism from~\cite{bahdanau2014neural} has achieved great success in several sequence modeling tasks, we integrate the attention mechanism into our HRNE model. We next introduce the attention mechanism part.

The core of the soft attention mechanism is that instead of just inputting the original sequence $(\mathbf{x}_1, \mathbf{x}_2, ..., \mathbf{x}_n)$ into a LSTM layer, dynamic weights are used to generate a new sequence $(\mathbf{v}_1, \mathbf{v}_2, ..., \mathbf{v}_m)$:
\begin{equation}
\mathbf{v}_t = \sum_{i=1}^n \alpha_i^{(t)} \mathbf{x}_i ,
\end{equation}
where $\sum_{i=1}^n \alpha_i^{(t)}=1$ and $\alpha_i^{(t)}$ will be calculated by an attention neural network  at each time step $t = 1, 2, \ldots, m$.

The attention weight $\alpha_i^{(t)}$ actually measures the relevance between the $i$-th element $\mathbf{x}_i$ of the input sequence  and the history information recorded by the LSTM $\mathbf{h}_{t-1}$. Hence a function is needed to calculate the relevance score:
\begin{equation}
e_i^{(t)}=w^\top \tanh(W_a  \mathbf{x}_i  + U_a \mathbf{h}_{t-1}+ b_a),
\end{equation}
where $w, W_a, U_a, b_a$ are all parameters and $\mathbf{h}_{t-1}$ is the hidden state of the LSTM at $(t{-1})$-th time step.

We need to calculate $e_i^{(t)}$ for $i = 1, 2, ..., n$ and then $\alpha_i^{(t)}$ could be calculated by:
\begin{equation}
\alpha_i^{(t)} = \exp(e_i^{(t)})/\sum_{j=1}^{n} \exp(e_j^{(t)}).
\end{equation}
The attention mechanism could make the LSTM pay attention to different temporal locations of the input sequence according to its backprop information, and when the input sequence and the output sequence are not aligned strictly, attention would especially be helpful.  We add attention units in three different positions in our video caption model: between the visual input and the LSTM filter, between the output of the filter and the second LSTM layer, between the output of our HRNE and the description decoder.

\subsection{Video Captioning}

 Our  HRNE  can be applied to several video processing tasks where feature vectors are required to represent videos. In this paper, we use video captioning, where temporal information plays an important role, to showcase the advantage of the proposed method.
 

We develop our video captioning model based on the general sequence to sequence model~\cite{sutskever2014sequence}, \ie, encoder-decoder framework, which is same as the previous works~\cite{yao2015describing,venugopalan2015sequence}. We use the general video encoder to map video sequences to feature vectors and then one-layer LSTM decoder conditioned on the video feature vector to generate description for the video. 

The overall objective function we are optimizing is the log-likelihood over the whole training set, 
\begin{equation}
\max_{\theta} \sum_{t = 1}^T \log \Pr(y_t | \mathbf{z}, y_{t-1}; \theta),
\end{equation}
where $y_t$ is a one-hot vector (1-of-N coding, where N is the size of the word vocabulary) used to represent the word at the $t$-th time step, $\mathbf{z}$ is the feature vector output by the video encoder and $\theta$ represents the video captioning model's parameters.

Similar to most recurrent neural network language models, we utilize a softmax layer to model the probability distribution of the next word over the word space, \ie, 
\begin{equation}
\Pr(y_t | \mathbf{z}, y_{t-1}; \theta) \propto \exp(y_t^\top W_y \mathbf{s}_t),
\label{eq:softmax}
\end{equation}

where 
\begin{equation}
\mathbf{s}_t = \tanh(W_z \mathbf{z} + W_h \mathbf{h}_t + W_e y_{t-1} + \mathbf{b}),
\label{eq:deepout}
\end{equation}
and $W_y, W_z, W_h, W_e$ and $b$ are all the parameters.

Eqn~(\ref{eq:deepout}) is an instance of deep output layer proposed in Pascanu~\etal~\cite{pascanu2013construct} and we find incorporating the deep output layer helps the model to converge faster and gets better performance. To make the model more robust, we adopt the Maxout~\cite{goodfellow2013maxout} scheme to calculate $\mathbf{s}_t$.

\section{Experimental Setup}
\label{sec:experiment_setup}
We utilize two standard video captioning benchmarks to validate the performance of our proposed method in the experiments:  the widely used Microsoft Video Description Corpus (MSVD)~\cite{chen2011collecting} and one recently proposed dataset the Montreal Video Annotation Dataset (M-VAD)~\cite{torabi2015using}.
\subsection{The Datasets}
\textbf{The Microsoft Video Description Corpus (MSVD)}: The Microsoft Video Description Corpus (MSVD)~\cite{chen2011collecting} contains 1,970 videos with multiple descriptions labeled by the Amazon Mechanical Turkers. Annotators are requested to provide a single sentence description to a picked up short clips. The total number of clip-description pairs is about 80,000. The original dataset consists of multi-lingual descriptions while we only focus on the English description as the previous works~\cite{venugopalan2014translating,venugopalan2015sequence,yao2015describing}. We utilize the standard splits provided in~\cite{venugopalan2014translating} for fair comparisons with state-of-the-art video captioning systems~\cite{venugopalan2014translating,venugopalan2015sequence,yao2015describing}, which separate the original dataset into training, validation and testing with 1,200 clips, 100 clips, and the remaining clips, respectively.

\textbf{The Montreal Video Annotation Dataset (M-VAD)}: The Montreal Video Annotation Dataset (M-VAD) is a newly collected large-scale video description dataset from the DVD descriptive video service (DVS) narrations. There are 92 DVD movies in the M-VAD dataset, which is further divided into 49,000 video clips. Each clip in the video has one corresponding narration as the groundtruth of the clip description. Since the narrations are generated in a semi-automatically transcribed way, the grammar used in the description is much more complicated than the one in MSVD. Same as previous works~\cite{yao2015describing,venugopalan2015sequence}, we utilize the standard splits provided in~\cite{torabi2015using}, which consists of 39,000 clips in the training set, 5,000 clips in the validation set, and 5,000 clips in the testing set. We note that the M-VAD dataset is much more challenging than the MSVD dataset and current state-of-the-art video description still produces very poor performance. 
\subsection{Preprocessing}

Visual Features: We use GoogLeNet~\cite{szegedy2014going} to extract the frame-level features in our experiment.  All the videos' lengths are kept to 160. For a video with more than  160 frames, we drop the extra frames. For a video without enough frames, we pad zero frames, following the common practices. Instead of directly inputting the features into HRNE, we learn a linear embedding of the features as the input of our model.

Description preprocessing: We convert all descriptions to lower case, and use the  PTBTokenizer in Stanford
CoreNLP tools\footnote{version 3.4.1}~\cite{manning2014stanford} to tokenize sentences and remove punctuation. This yields a vocabulary of 12,976 in size for the MSVD dataset and a vocabulary of 15,567 in size for the M-VAD dataset.

\subsection{Evaluation Metrics}
Several standard metrics such as BLEU~\cite{papineni2002bleu}, METEOR~\cite{denkowski2014meteor}, ROUGE-L~\cite{lin2004rouge} and CIDEr~\cite{vedantam2014cider} are used commonly for evaluating visual captioning tasks, mainly following the machine translation field. The authors of \cite{vedantam2014cider} evaluated the above four metrics in terms of the consistency with human judgment, and found that METEOR is always better than BLEU and ROUGE. Thus, \emph{METEOR is used as the main metric in the evaluation}. We utilize the Microsoft COCO evaluation server~\cite{chen2015microsoft} to obtain all the results reported in this paper, which makes our results directly comparable with the previous works.
\subsection{Compared Algorithms}

\begin{itemize}
\item{FGM~\cite{thomason2014integrating}:} It first obtains confidences on subject, verb, object and scene elements. Then a factor graph model is used to infer the most likely (subject, verb, object) tuple in the video. Finally it generates sentence based on a template.

\item{Average pooling + LSTM decoder~\cite{venugopalan2014translating} (denoted as Mean pool):}
The average pooling frame-level features is treated as the representation of the whole video. LSTM is  utilized as  a recurrent language model to produce the description given the visual feature.

\item{S2VT~\cite{venugopalan2015sequence}:} This is an encoder-decoder model. It consists of two phases. In the first phase, it serves as a video encoder and in the second phase, it stops accepting video sequence and begins generating video descriptions.

\item{Temporal Attention~\cite{yao2015describing} (SA):} It applies attention mechanism on temporal locations and then utilizes the recurrent language model LSTM to generate the video description.

\item{LSTM embdding~\cite{pan2015jointly} (LSTM-E):} It uses embedding layers to project the visual feature and text feature into one space, with a modified  loss between description and visual features.

\item{Paragraph RNN decoder~\cite{yu2015video} (p-RNN):} It introduces a hierarchical structure in \emph{decoder} for language processing and introduce the paragraph description in addition to the standard sentence description.
\end{itemize}

\subsection{Training Details}
In the training phase, we add a begin-of-sentence tag $<$BOS$>$ to start each sentence and an end-of-sentence tag $<$EOS$>$ to end each sentence, so that our captioning model can deal with sentences of varying lengths. In the testing phase, we input $<$BOS$>$ into video decoder to start generating video descriptions and during each step, we choose the word with the maximun probability after softmax until we reach $<$EOS$>$.

We adopt different parameter settings to train different datasets. When we are training on MSVD, we use the following settings: All the LSTM units are set to 1,024, the visual feature embedding size and the word embedding size are set as 512 empirically. 
When training on M-VAD, we find our HRNE is easier to overfit than in MSVD, so we set all the LSTM units to be 512 and still keep the visual feature embedding size and the word embedding size to be half of the number of LSTM units. As the videos in the two datasets are very short,  a two-layer HRNE is sufficient to capture the temporal structure of videos. Nevertheless, one may use HRNE with more layers to deal with longer videos. 

The length of the LSTM chain at the bottom layer is 8, and we set the stride to be 8 in all the experiments.  We set the size of mini-batch as 128. We apply the first-order optimizer ADAM to minimize the negative log-likelihood loss for the training process and  we set the learning rate $\eta = 2\e{-4}$, the decay parameters $\beta_1=0.9$, $\beta_2=0.999$ as defaulted in Kingman and Ba~\cite{kingma2014adam}, which generally shows good performance and does not need heavily tuned. Since we observe serious overfitting problems when training our model on M-VAD dataset, we apply the simple yet effective neural model regularization method Dropout~\cite{srivastava2014dropout} with rate of 0.5 on the input and the output of LSTMs but not on the recurrent transitions as suggested by Zaremba~\etal ~\cite{zaremba2014recurrent}. We find that the proposed model has better generalization ability in this way, empirically.

We train the model for 200 epoches, or stop the training until  the evaluation metric does not improve on the validation set.  We utilize Theano~\cite{Bastien-Theano-2012,bergstra+al:2010-scipy} framework to conduct our experiments.
\section{Experimental Results}
\label{sec:results}
We evaluate our HRNE model on video captioning on both MSVD and M-VAD. We report results on MSVD in Table \ref{tb:MSVD_RGB} and Table \ref{tb:MSVD}. Note that \emph{only RGB GoogLeNet features are adopted by our HRNE} in all the  experiments. To make a fair comparison, we firstly report the results only using static frame-level features  in Table \ref{tb:MSVD_RGB}. We additionally compare our HRNE with only one ConvNet feature to other video captioning systems which combine multiple ConvNet features in Table \ref{tb:MSVD}. Lastly, we conduct the experiment on the more challenging dataset M-VAD, and report the results in Table \ref{tb:MVAD} .

\subsection{Experiment results on the MSVD dataset}

We report experiment results where only static frame-level features are used in Table \ref{tb:MSVD_RGB} on the MSVD dataset. Both Mean pool and SA  ignore temporal dependencies along video sequences. They adopt the weighted averages of frame-level features to represent videos. Our HRNE outperforms both Mean pool and SA, due to the  exploration of temporal information of videos. Hierarchical description decoder is adopted in p-RNN to generate complex descriptions, while our HRNE has better performance than p-RNN, which indicates exploring temporal information of videos is more important for video captioning. S2VT uses one-layer LSTM as video encoder to explore videos' temporal information. Our HRNE achieves better result than S2VT, because the hierarchical structure in HRNE  reduces the length of input flow and composites multiple consecutive input at a higher level, which increases the learning capability and enables our model encode richer temporal information of multiple granularities. To further improve our HRNE, we add attention mechanism, which again improves its performance.

We additionally compare our HRNE to other video captioning systems with fusion in Table \ref{tb:MSVD}. In this experiment, our HRNE only uses one ConvNet feature as input but the compared systems combine multiple ConvNet features. Mean pool model  achieves the best result on AlexNet with the model pre-trained on COCO \cite{lin2014microsoft} and it achieves 29.1\% METEOR. S2VT achieves the best result with RGB frames on VGGNet and optical flows on AlexNet and it achieves 29.8\% on METEOR. SA model gets the best result with GoogLeNet and 3D-CNN. It achieves 29.6\% in METEOR and 41.9\% in BLEU-4. Our HRNE achieves the best result in METEOR. It means although adding more features helps improve  video captioning systems' performance,  our method still achieves the best performance. This result confirms the effectiveness of our HRNE. We notice that p-RNN outperforms our HRNE in terms of BLEU. However, our method  outperforms p-RNN in almost all other cases (see Table \ref{tb:MSVD_RGB} and Table \ref{tb:MSVD}) and, more importantly, as demonstrated in \cite{vedantam2014cider},  METEOR is more reliable than BLEU.

In  Table \ref{tb:caption}, we show a few examples of the descriptions generated by our method. We notice that our HRNE can generate an accurate description of the video even in some difficult cases. In addition, the results with the attention mechanism is generally better than those without the attention mechanism, which is consistent with the results reported in Table \ref{tb:MSVD_RGB} and Table \ref{tb:MSVD}.


\begin{table}
\footnotesize
\begin{center}
\begin{tabular}{c|c||c|c|c|c}
\hline
Model & METEOR & B@1 & B@2 & B@3 & B@4 \\
\hline
FGM~\cite{thomason2014integrating}&   23.9 & - & - & - & - \\
Mean pool~\cite{yao2015describing} & 28.7 & - & - & - & 38.7 \\
SA~\cite{yao2015describing} & 29.0  &  - & - & - & 40.3  \\
S2VT~\cite{venugopalan2015sequence} & 29.2 & - & - & - & - \\
LSTM-E~\cite{pan2015jointly} &29.5  & 74.9 & 60.9 & 50.6 & 40.2 \\
p-RNN~\cite{yu2015video} & 31.1 & 77.3 & 64.5 & 54.6 & \bf{44.3} \\
\hline
\hline
HRNE  & 32.1 & 78.4 & 66.1 & 55.1 & 43.6 \\
HRNE with attention & \bf{33.1} & \bf{79.2} & \bf{66.3} & \bf{55.1} & 43.8\\
\hline
\end{tabular}
\caption{Experiment results on the MSVD dataset. We compare our method with the baselines using \emph{static frame-level features only} in this table.}
\label{tb:MSVD_RGB}
\end{center}
\vspace{-0.5cm}
\end{table}

\begin{table}
\scriptsize
\begin{center}
\begin{tabular}{c|c||c|c|c|c}
\hline
Model & METEOR & B@1 & B@2 & B@3 & B@4 \\
\hline
LSTM-E-(A) ~\cite{pan2015jointly} & 28.3 & 74.5 & 59.8 & 49.3 & 38.9 \\
LSTM-E-(V) ~\cite{pan2015jointly} & 29.5 & 74.9 & 60.9 & 50.6 & 40.2 \\
LSTM-E-(C) ~\cite{pan2015jointly} & 29.9 & 75.7 & 62.3 & 52.0 & 41.7 \\
LSTM-E-(V)+(C) ~\cite{pan2015jointly} & 31.0 & 78.8 & 66.0 & 55.4 & 45.3 \\
\hline
p-RNN-(V)~\cite{yu2015video} & 31.1 & 77.3 & 64.5 & 54.6 & 44.3 \\
p-RNN-(C)~\cite{yu2015video} & 30.3 & 79.7 & 67.9 & 57.9 & 47.4 \\
p-RNN-(V)+(C)~\cite{yu2015video} & 32.6 & \bf{81.5} & \bf{70.4} & \bf{60.4} & \bf{49.9} \\
\hline
\hline
HRNE-(G) & 32.1 & 78.4 & 66.1 & 55.1 & 43.6 \\
HRNE with attention-(G) & \bf{33.1} & 79.2 & 66.3 & 55.1 & 43.8\\
\hline
\end{tabular}
\caption{Experiment results on the MSVD dataset with fusion. (A) denotes AlexNet, (V) denotes VGGNet, (C) denotes C3D and (G) denotes GoogLeNet in the model's name. Note that our method uses (G) only without fusion.}
\label{tb:MSVD}
\end{center}
\vspace{-0.5cm}
\end{table}

\subsection{Experiment results on the M-VAD dataset}
Table \ref{tb:MVAD} reports the results on M-VAD. Compared with MSVD, M-VAD is a more challenging dataset, because it contains more visual concepts and complex sentence structures. Since the result on BLEU metric is close to 0\footnote{SA~\cite{yao2015describing} achieves only 0.7\% BLEU-4 on this dataset.}, we do not consider BLEU metric in this experiment. Our HRNE achieves 5.8\% in METEOR, which outperforms both S2VT and SA\footnote{Only S2VT and SA have reported result on this challenging dataset.}. After adding the attention mechanism, our performance (in METEOR) is further improved from 5.8\%  to 6.8\%. Such performance even outperforms S2VT which combines M-VAD and MPII-MD \cite{rohrbach15cvpr} for training. Because combining two datasets introduces much more training data than just one dataset as the standard setting we used for training, this result again validates the effectiveness of our HRNE.

\begin{table*}
\scriptsize
\begin{center}
\begin{tabular}{c c c}
\includegraphics[width=0.15\textwidth,height=0.088\textheight]{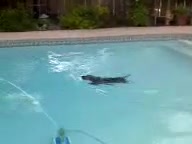}  \includegraphics[width=0.15\textwidth,height=0.088\textheight]{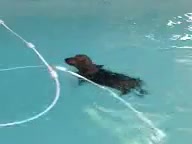} 
& \includegraphics[width=0.15\textwidth,height=0.088\textheight]{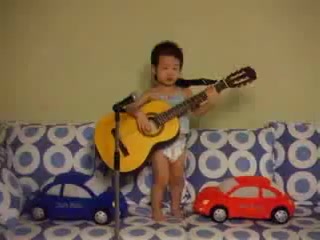}  \includegraphics[width=0.15\textwidth,height=0.088\textheight]{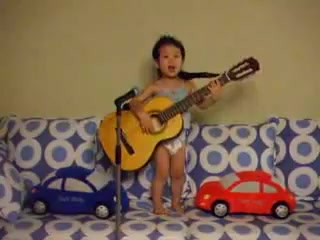}&
\includegraphics[width=0.15\textwidth,height=0.088\textheight]{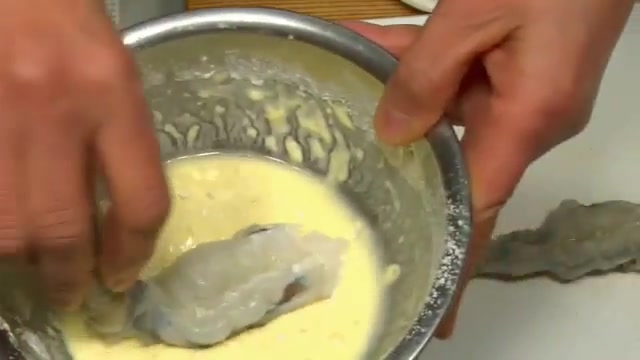}  \includegraphics[width=0.15\textwidth,height=0.088\textheight]{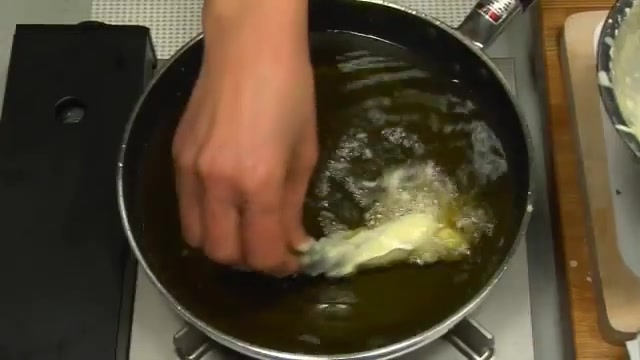} \\
{\bf HRNE}: A man is swimming in the water. &  {\bf HRNE}： A man is playing a guitar & {\bf HRNE}: A woman is adding noodles into a pot\\
{\bf HRNE with attention}: A dog is swimming. & {\bf HRNE with attention}: A man is playing a guitar. & {\bf HRNE with attention}: A woman is cooking. \\
{\bf Ground truth}: A dog is swimming in a pool. & {\bf Ground truth}:  A boy is playing a guitar. & {\bf Ground truth}: A woman dips a shrimp in batter.\\
\\
\includegraphics[width=0.15\textwidth,height=0.088\textheight]{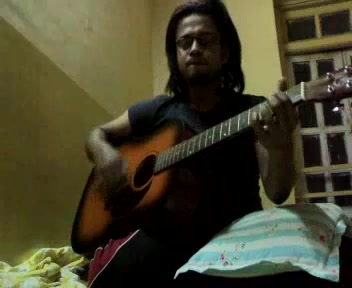}  \includegraphics[width=0.15\textwidth,height=0.088\textheight]{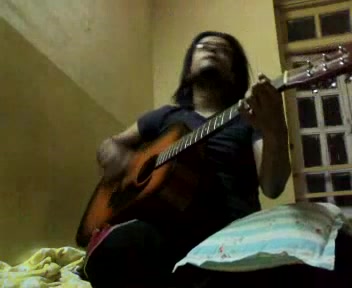} 
& \includegraphics[width=0.15\textwidth,height=0.088\textheight]{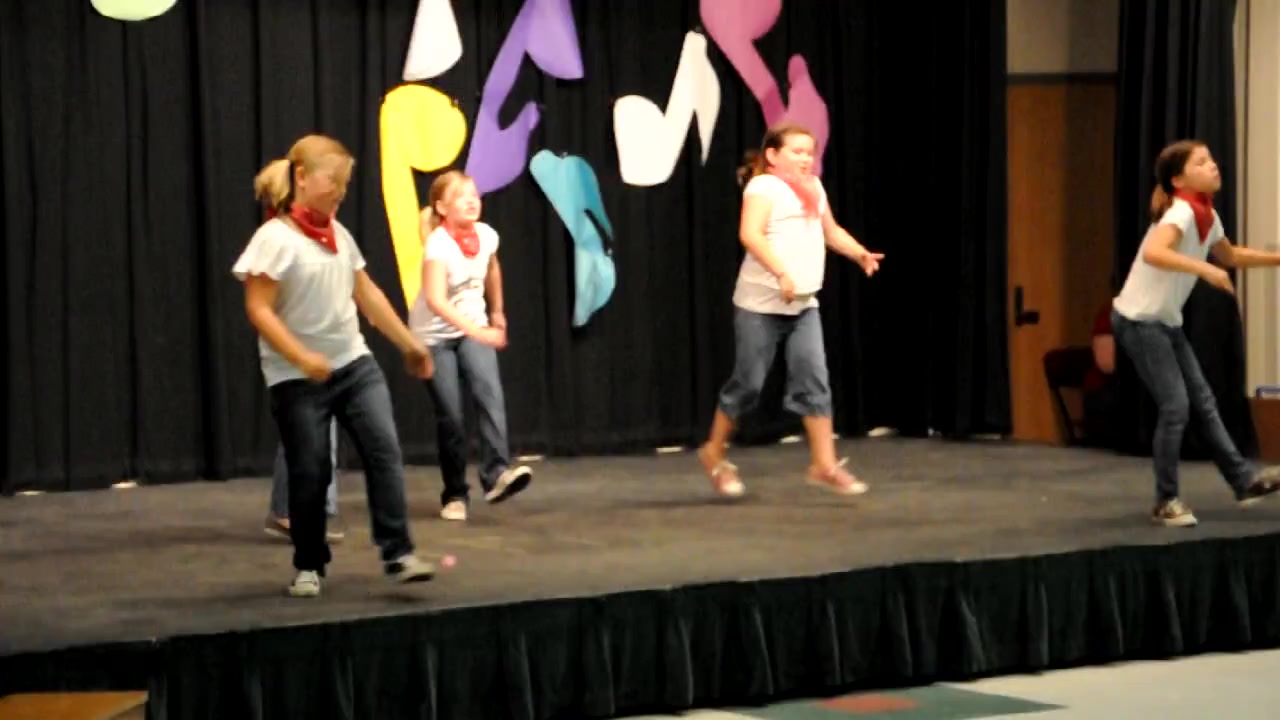}  \includegraphics[width=0.15\textwidth,height=0.088\textheight]{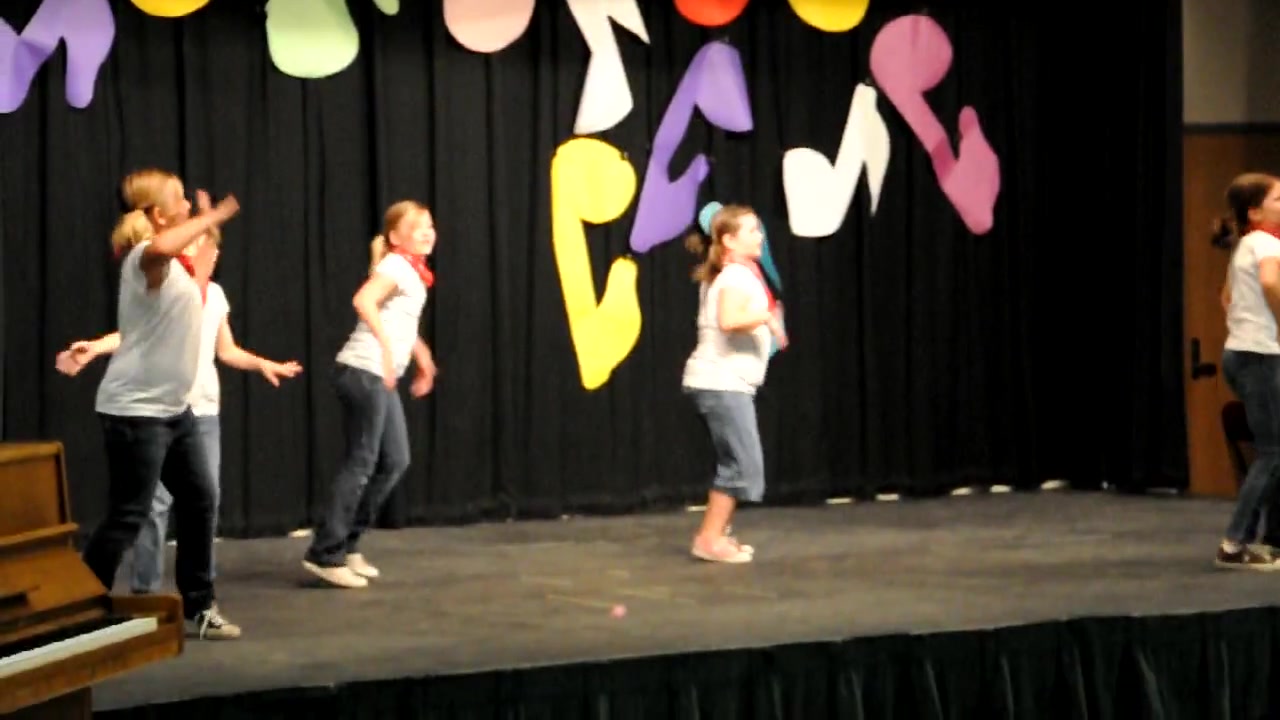}&
\includegraphics[width=0.15\textwidth,height=0.088\textheight]{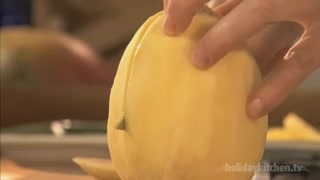}  \includegraphics[width=0.15\textwidth,height=0.088\textheight]{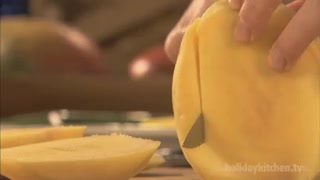} \\

{\bf HRNE}: A man is playing a guitar. & {\bf HRNE}: A group of people are dancing. &  {\bf HRNE}: A person is preparing an egg.\\
{\bf HRNE with attention}:  A man is playing a guitar & {\bf HRNE with attention}: A group of people are dancing & {\bf HRNE with attention}: A woman is peeling a mango.\\
{\bf Ground truth}:  A man plays a guitar. & {\bf Ground truth}: A group of young girls are dancing on stage. & {\bf Ground truth}:  A mango is being sliced. \\
\\
 
\includegraphics[width=0.15\textwidth,height=0.088\textheight]{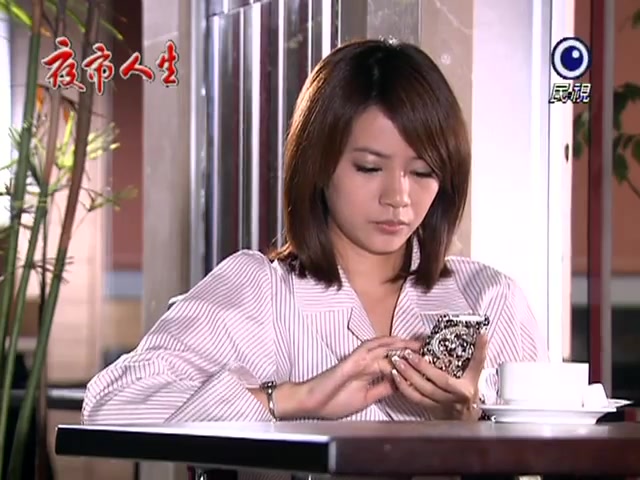}  \includegraphics[width=0.15\textwidth,height=0.088\textheight]{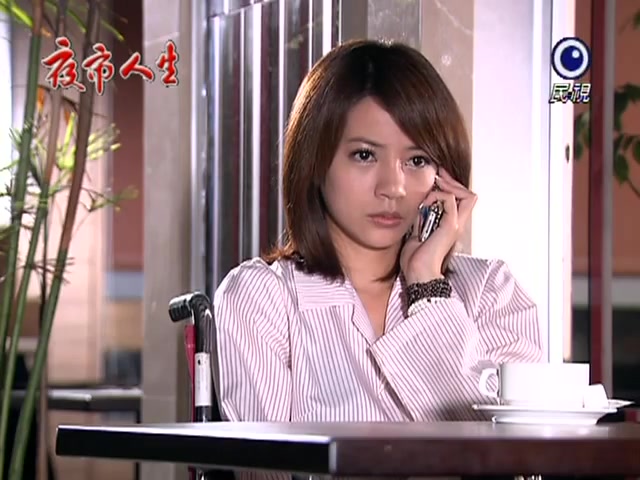} 
& \includegraphics[width=0.15\textwidth,height=0.088\textheight]{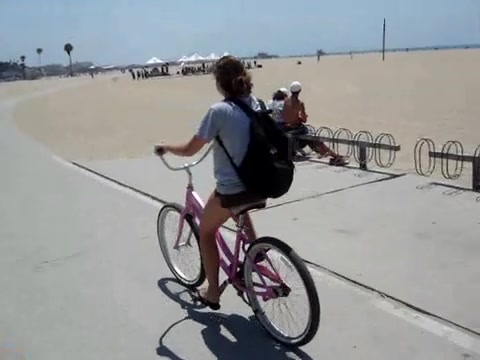}  \includegraphics[width=0.15\textwidth,height=0.088\textheight]{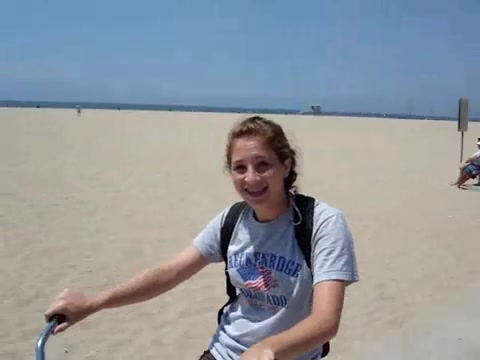}&
\includegraphics[width=0.15\textwidth,height=0.088\textheight]{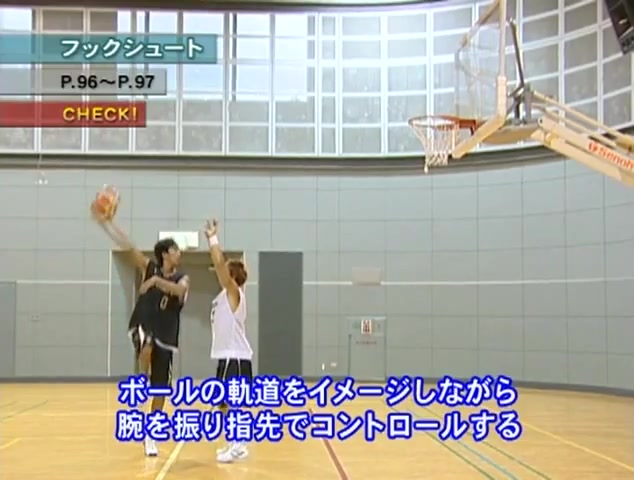}  \includegraphics[width=0.15\textwidth,height=0.088\textheight]{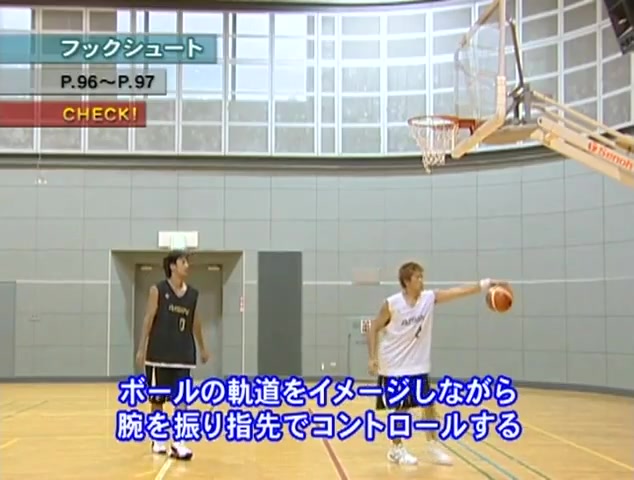} \\

{\bf HRNE}:  A girl is talking. & {\bf HRNE}: A man is riding a bike. &  {\bf HRNE}: A man is doing a machine.\\
{\bf HRNE with attention}:  A young girl is talking on a telephone & {\bf HRNE with attention}: A man is riding a bike & {\bf HRNE with attention}: A man is doing a dance.\\
{\bf Ground truth}:  A woman dials a cell phone. & {\bf Ground truth}: A biker rides along the beach. & {\bf Ground truth}: A basketball player is doing a hook shot. \\
 
\end{tabular}
\caption{Example results on the MSVD Youtube video dataset. We present the video descriptions generated by our HRNE.}
\label{tb:caption}
\end{center}
\end{table*}

\begin{table}[t]
\footnotesize
\begin{center}
\begin{tabular}{l|c}
\hline
Model & METEOR \\
\hline
SA-GoogleNet+3D-CNN \cite{yao2015describing} &  5.7 \\
SA-GoogleNet+3D-CNN \cite{yao2015describing}\footnotemark & 4.1  \\
S2VT-RGB(VGG)  \cite{venugopalan2015sequence} & 5.6 \\
\hline
HRNE & 5.8 \\
HRNE (with attention) & \bf{6.8} \\
\hline
\end{tabular}
\caption{Experiment results on the M-VAD dataset.}
\label{tb:MVAD}
\end{center}
\end{table}
\footnotetext{\cite{venugopalan2015sequence} notes that \cite{yao2015describing} achieves 4.1\% METEOR with the same evaluation script as~\cite{venugopalan2015sequence}, while the 5.7\% METEOR reported in~\cite{yao2015describing} is caused by different tokenization.}

\section{Conclusions and Future Work}
\label{sec:conclusion}
In this paper, we proposed a new method, namely Hierarchical Recurrent Neural Encoder (HRNE), to generate video representation with emphasis on temporal modeling. Compared to existing approaches, the proposed HRNE is more capable of video modeling because 1) HRNE reduces the length of input information flow and exploits temporal structure in longer range at a higher level; 2) more non-linearity and flexibility are added in HRNE; and 3) HRNE exploits temporal transitions with multiple granularities. Extensive experiments in video captioning demonstrate the efficacy of HRNE.

 Last but not least, the proposed video representation is generic which can be applied to a wide range of video analysis applications.  We will explore the application of the encoder on video classification in the future work, which plugs with a softmax classifier upon the encoder and video labels instead of the LSTM language decoder in this work to validate the generalization capability of this framework.

{\small
\bibliographystyle{ieee}
\bibliography{HRNE}

\begin{thebibliography}{10}\itemsep=-1pt

\bibitem{bahdanau2014neural}
D.~Bahdanau, K.~Cho, and Y.~Bengio.
\newblock Neural machine translation by jointly learning to align and
  translate.
\newblock In {\em ICLR}, 2015.

\bibitem{Bastien-Theano-2012}
F.~Bastien, P.~Lamblin, R.~Pascanu, J.~Bergstra, I.~J. Goodfellow, A.~Bergeron,
  N.~Bouchard, and Y.~Bengio.
\newblock Theano: new features and speed improvements.
\newblock NIPS Workshop, 2012.

\bibitem{bengio1994learning}
Y.~Bengio, P.~Simard, and P.~Frasconi.
\newblock Learning long-term dependencies with gradient descent is difficult.
\newblock {\em Neural Networks, IEEE Transactions on}, 5(2):157--166, 1994.

\bibitem{bergstra+al:2010-scipy}
J.~Bergstra, O.~Breuleux, F.~Bastien, P.~Lamblin, R.~Pascanu, G.~Desjardins,
  J.~Turian, D.~Warde-Farley, and Y.~Bengio.
\newblock Theano: a {CPU} and {GPU} math expression compiler.
\newblock In {\em SciPy)}, June 2010.

\bibitem{chen2011collecting}
D.~L. Chen and W.~B. Dolan.
\newblock Collecting highly parallel data for paraphrase evaluation.
\newblock In {\em ACL}, 2011.

\bibitem{chen2015microsoft}
X.~Chen, H.~Fang, T.-Y. Lin, R.~Vedantam, S.~Gupta, P.~Dollar, and C.~L.
  Zitnick.
\newblock Microsoft {COCO} captions: Data collection and evaluation server.
\newblock {\em arXiv preprint arXiv:1504.00325}, 2015.

\bibitem{cho2015learning}
K.~Cho, B.~Van~Merri{\"e}nboer, C.~Gulcehre, D.~Bahdanau, F.~Bougares,
  H.~Schwenk, and Y.~Bengio.
\newblock Learning phrase representations using {RNN} encoder-decoder for
  statistical machine translation.
\newblock In {\em EMNLP}, 2015.

\bibitem{denkowski2014meteor}
M.~Denkowski and A.~Lavie.
\newblock Meteor universal: Language specific translation evaluation for any
  target language.
\newblock In {\em EACL}, 2014.

\bibitem{donahue2014long}
J.~Donahue, L.~A. Hendricks, S.~Guadarrama, M.~Rohrbach, S.~Venugopalan,
  K.~Saenko, and T.~Darrell.
\newblock Long-term recurrent convolutional networks for visual recognition and
  description.
\newblock In {\em CVPR}, 2015.

\bibitem{goodfellow2013maxout}
I.~Goodfellow, D.~Warde-farley, M.~Mirza, A.~Courville, and Y.~Bengio.
\newblock Maxout networks.
\newblock In {\em ICML}, 2013.

\bibitem{hochreiter1997long}
S.~Hochreiter and J.~Schmidhuber.
\newblock Long short-term memory.
\newblock {\em Neural computation}, 9(8):1735--1780, 1997.

\bibitem{jegou2010aggregating}
H.~J{\'e}gou, M.~Douze, C.~Schmid, and P.~P{\'e}rez.
\newblock Aggregating local descriptors into a compact image representation.
\newblock In {\em CVPR}, 2010.

\bibitem{ji20133d}
S.~Ji, W.~Xu, M.~Yang, and K.~Yu.
\newblock 3d convolutional neural networks for human action recognition.
\newblock {\em TPAMI}, 35(1):221--231, 2013.

\bibitem{karpathy2014large}
A.~Karpathy, G.~Toderici, S.~Shetty, T.~Leung, R.~Sukthankar, and L.~Fei-Fei.
\newblock Large-scale video classification with convolutional neural networks.
\newblock In {\em CVPR}, 2014.

\bibitem{kingma2014adam}
D.~Kingma and J.~Ba.
\newblock {ADAM}: A method for stochastic optimization.
\newblock In {\em ICLR}, 2015.

\bibitem{krizhevsky2012imagenet}
A.~Krizhevsky, I.~Sutskever, and G.~E. Hinton.
\newblock {ImageNet} classification with deep convolutional neural networks.
\newblock In {\em NIPS}, 2012.

\bibitem{lin2004rouge}
C.-Y. Lin.
\newblock {ROUGE}: A package for automatic evaluation of summaries.
\newblock In {\em ACL workshop}, 2004.

\bibitem{lin2014microsoft}
T.-Y. Lin, M.~Maire, S.~Belongie, J.~Hays, P.~Perona, D.~Ramanan,
  P.~Doll{\'a}r, and C.~L. Zitnick.
\newblock Microsoft {COCO}: Common objects in context.
\newblock In {\em ECCV}. 2014.

\bibitem{manning2014stanford}
C.~D. Manning, M.~Surdeanu, J.~Bauer, J.~Finkel, S.~J. Bethard, and
  D.~McClosky.
\newblock The {Stanford} {CoreNLP} natural language processing toolkit.
\newblock In {\em ACL}, 2014.

\bibitem{ng2015beyond}
J.~Y.-H. Ng, M.~Hausknecht, S.~Vijayanarasimhan, O.~Vinyals, R.~Monga, and
  G.~Toderici.
\newblock Beyond short snippets: Deep networks for video classification.
\newblock In {\em CVPR}, 2015.

\bibitem{pan2015jointly}
Y.~Pan, T.~Mei, T.~Yao, H.~Li, and Y.~Rui.
\newblock Jointly modeling embedding and translation to bridge video and
  language.
\newblock {\em arXiv preprint arXiv:1505.01861}, 2015.

\bibitem{papineni2002bleu}
K.~Papineni, S.~Roukos, T.~Ward, and W.-J. Zhu.
\newblock {BLEU}: a method for automatic evaluation of machine translation.
\newblock In {\em ACL}, 2002.

\bibitem{pascanu2013construct}
R.~Pascanu, C.~Gulcehre, K.~Cho, and Y.~Bengio.
\newblock How to construct deep recurrent neural networks.
\newblock {\em arXiv preprint arXiv:1312.6026}, 2013.

\bibitem{rohrbach15cvpr}
A.~Rohrbach, M.~Rohrbach, N.~Tandon, and B.~Schiele.
\newblock A dataset for movie description.
\newblock In {\em CVPR}, 2015.

\bibitem{sanchez2013image}
J.~S{\'a}nchez, F.~Perronnin, T.~Mensink, and J.~Verbeek.
\newblock Image classification with the fisher vector: Theory and practice.
\newblock {\em IJCV}, 105(3):222--245, 2013.

\bibitem{simonyan2014two}
K.~Simonyan and A.~Zisserman.
\newblock Two-stream convolutional networks for action recognition in videos.
\newblock In {\em NIPS}, 2014.

\bibitem{Simonyan14c}
K.~Simonyan and A.~Zisserman.
\newblock Very deep convolutional networks for large-scale image recognition.
\newblock In {\em ICLR}, 2015.

\bibitem{sivic2003video}
J.~Sivic and A.~Zisserman.
\newblock Video google: A text retrieval approach to object matching in videos.
\newblock In {\em CVPR}, 2003.

\bibitem{Sordoni2015hierarchical}
A.~Sordoni, Y.~Bengio, H.~Vahabi, C.~Lioma, J.~G. Simonsen, and J.-Y. Nie.
\newblock A hierarchical recurrent encoder-decoder for generative context-aware
  query suggestion.
\newblock In {\em CIKM}, 2015.

\bibitem{srivastava2014dropout}
N.~Srivastava, G.~Hinton, A.~Krizhevsky, I.~Sutskever, and R.~Salakhutdinov.
\newblock Dropout: A simple way to prevent neural networks from overfitting.
\newblock {\em JMLR}, 15(1):1929--1958, 2014.

\bibitem{sutskever2014sequence}
I.~Sutskever, O.~Vinyals, and Q.~V. Le.
\newblock Sequence to sequence learning with neural networks.
\newblock In {\em NIPS}, 2014.

\bibitem{szegedy2014going}
C.~Szegedy, W.~Liu, Y.~Jia, P.~Sermanet, S.~Reed, D.~Anguelov, D.~Erhan,
  V.~Vanhoucke, and A.~Rabinovich.
\newblock Going deeper with convolutions.
\newblock In {\em CVPR}, 2015.

\bibitem{thomason2014integrating}
J.~Thomason, S.~Venugopalan, S.~Guadarrama, K.~Saenko, and R.~Mooney.
\newblock Integrating language and vision to generate natural language
  descriptions of videos in the wild.
\newblock In {\em COLING}, 2014.

\bibitem{torabi2015using}
A.~Torabi, C.~Pal, H.~Larochelle, and A.~Courville.
\newblock Using descriptive video services to create a large data source for
  video annotation research.
\newblock {\em arXiv preprint arXiv:1503.01070}, 2015.

\bibitem{DBLP:journals/corr/TranBFTP14}
D.~Tran, L.~D. Bourdev, R.~Fergus, L.~Torresani, and M.~Paluri.
\newblock {C3D:} generic features for video analysis.
\newblock In {\em ICCV}, 2015.

\bibitem{vedantam2014cider}
R.~Vedantam, C.~L. Zitnick, and D.~Parikh.
\newblock {CIDEr}: Consensus-based image description evaluation.
\newblock In {\em CVPR}, 2015.

\bibitem{venugopalan2015sequence}
S.~Venugopalan, M.~Rohrbach, J.~Donahue, R.~J. Mooney, T.~Darrell, and
  K.~Saenko.
\newblock Sequence to sequence - video to text.
\newblock {\em arXiv preprint arXiv:1505.00487v2}.

\bibitem{venugopalan2014translating}
S.~Venugopalan, H.~Xu, J.~Donahue, M.~Rohrbach, R.~Mooney, and K.~Saenko.
\newblock Translating videos to natural language using deep recurrent neural
  networks.
\newblock In {\em NAACL-HLT}, 2015.

\bibitem{wang2011action}
H.~Wang, A.~Kl{\"a}ser, C.~Schmid, and C.-L. Liu.
\newblock Action recognition by dense trajectories.
\newblock In {\em CVPR}, 2011.

\bibitem{wang2013action}
H.~Wang and C.~Schmid.
\newblock Action recognition with improved trajectories.
\newblock In {\em ICCV}, 2013.

\bibitem{xu2015discriminative}
Z.~Xu, Y.~Yang, and A.~G. Hauptmann.
\newblock A discriminative {CNN} video representation for event detection.
\newblock In {\em CVPR}, 2015.

\bibitem{yao2015describing}
L.~Yao, A.~Torabi, K.~Cho, N.~Ballas, C.~Pal, H.~Larochelle, and A.~Courville.
\newblock Describing videos by exploiting temporal structure.
\newblock In {\em ICCV}, 2015.

\bibitem{yu2015video}
H.~Yu, J.~Wang, Z.~Huang, Y.~Yang, and W.~Xu.
\newblock Video paragraph captioning using hierarchical recurrent neural
  networks.
\newblock {\em arXiv preprint arXiv:1510.07712}, 2015.

\bibitem{DBLP:journals/corr/ZarembaS14}
W.~Zaremba and I.~Sutskever.
\newblock Learning to execute.
\newblock In {\em ICLR}, 2015.

\bibitem{zaremba2014recurrent}
W.~Zaremba, I.~Sutskever, and O.~Vinyals.
\newblock Recurrent neural network regularization.
\newblock {\em arXiv preprint arXiv:1409.2329}, 2014.

\end{thebibliography}


\begin{thebibliography}{10}\itemsep=-1pt

\bibitem{bahdanau2014neural}
D.~Bahdanau, K.~Cho, and Y.~Bengio.
\newblock Neural machine translation by jointly learning to align and
  translate.
\newblock {\em arXiv preprint arXiv:1409.0473}, 2014.

\bibitem{bengio1994learning}
Y.~Bengio, P.~Simard, and P.~Frasconi.
\newblock Learning long-term dependencies with gradient descent is difficult.
\newblock {\em Neural Networks, IEEE Transactions on}, 5(2):157--166, 1994.

\bibitem{chen2011collecting}
D.~L. Chen and W.~B. Dolan.
\newblock Collecting highly parallel data for paraphrase evaluation.
\newblock In {\em Proceedings of the 49th Annual Meeting of the Association for
  Computational Linguistics: Human Language Technologies-Volume 1}, pages
  190--200. Association for Computational Linguistics, 2011.

\bibitem{cho2014properties}
K.~Cho, B.~van Merri{\"e}nboer, D.~Bahdanau, and Y.~Bengio.
\newblock On the properties of neural machine translation: Encoder-decoder
  approaches.
\newblock {\em arXiv preprint arXiv:1409.1259}, 2014.

\bibitem{cho2014learning}
K.~Cho, B.~Van~Merri{\"e}nboer, C.~Gulcehre, D.~Bahdanau, F.~Bougares,
  H.~Schwenk, and Y.~Bengio.
\newblock Learning phrase representations using rnn encoder-decoder for
  statistical machine translation.
\newblock {\em arXiv preprint arXiv:1406.1078}, 2014.

\bibitem{donahue2014long}
J.~Donahue, L.~A. Hendricks, S.~Guadarrama, M.~Rohrbach, S.~Venugopalan,
  K.~Saenko, and T.~Darrell.
\newblock Long-term recurrent convolutional networks for visual recognition and
  description.
\newblock {\em arXiv preprint arXiv:1411.4389}, 2014.

\bibitem{el1995hierarchical}
S.~El~Hihi and Y.~Bengio.
\newblock Hierarchical recurrent neural networks for long-term dependencies.
\newblock In {\em NIPS}, pages 493--499. Citeseer, 1995.

\bibitem{graves2014towards}
A.~Graves and N.~Jaitly.
\newblock Towards end-to-end speech recognition with recurrent neural networks.
\newblock In {\em Proceedings of the 31st International Conference on Machine
  Learning (ICML-14)}, pages 1764--1772, 2014.

\bibitem{hochreiter1997long}
S.~Hochreiter and J.~Schmidhuber.
\newblock Long short-term memory.
\newblock {\em Neural computation}, 9(8):1735--1780, 1997.

\bibitem{ji20133d}
S.~Ji, W.~Xu, M.~Yang, and K.~Yu.
\newblock 3d convolutional neural networks for human action recognition.
\newblock {\em Pattern Analysis and Machine Intelligence, IEEE Transactions
  on}, 35(1):221--231, 2013.

\bibitem{karpathy2014deep}
A.~Karpathy and L.~Fei-Fei.
\newblock Deep visual-semantic alignments for generating image descriptions.
\newblock {\em arXiv preprint arXiv:1412.2306}, 2014.

\bibitem{karpathy2014large}
A.~Karpathy, G.~Toderici, S.~Shetty, T.~Leung, R.~Sukthankar, and L.~Fei-Fei.
\newblock Large-scale video classification with convolutional neural networks.
\newblock In {\em Computer Vision and Pattern Recognition (CVPR), 2014 IEEE
  Conference on}, pages 1725--1732. IEEE, 2014.

\bibitem{krizhevsky2012imagenet}
A.~Krizhevsky, I.~Sutskever, and G.~E. Hinton.
\newblock Imagenet classification with deep convolutional neural networks.
\newblock In {\em Advances in neural information processing systems}, pages
  1097--1105, 2012.

\bibitem{lecun1998gradient}
Y.~LeCun, L.~Bottou, Y.~Bengio, and P.~Haffner.
\newblock Gradient-based learning applied to document recognition.
\newblock {\em Proceedings of the IEEE}, 86(11):2278--2324, 1998.

\bibitem{mao2014explain}
J.~Mao, W.~Xu, Y.~Yang, J.~Wang, and A.~L. Yuille.
\newblock Explain images with multimodal recurrent neural networks.
\newblock {\em arXiv preprint arXiv:1410.1090}, 2014.

\bibitem{ng2015beyond}
J.~Y.-H. Ng, M.~Hausknecht, S.~Vijayanarasimhan, O.~Vinyals, R.~Monga, and
  G.~Toderici.
\newblock Beyond short snippets: Deep networks for video classification.
\newblock {\em arXiv preprint arXiv:1503.08909}, 2015.

\bibitem{DBLP:journals/corr/PanMYLR15}
Y.~Pan, T.~Mei, T.~Yao, H.~Li, and Y.~Rui.
\newblock Jointly modeling embedding and translation to bridge video and
  language.
\newblock {\em CoRR}, abs/1505.01861, 2015.

\bibitem{rohrbach2015long}
A.~Rohrbach, M.~Rohrbach, and B.~Schiele.
\newblock The long-short story of movie description.
\newblock In {\em Pattern Recognition}, pages 209--221. Springer, 2015.

\bibitem{serban2015hierarchical}
I.~V. Serban, A.~Sordoni, Y.~Bengio, A.~Courville, and J.~Pineau.
\newblock Hierarchical neural network generative models for movie dialogues.
\newblock {\em arXiv preprint arXiv:1507.04808}, 2015.

\bibitem{simonyan2014two}
K.~Simonyan and A.~Zisserman.
\newblock Two-stream convolutional networks for action recognition in videos.
\newblock In {\em Advances in Neural Information Processing Systems}, pages
  568--576, 2014.

\bibitem{Simonyan14c}
K.~Simonyan and A.~Zisserman.
\newblock Very deep convolutional networks for large-scale image recognition.
\newblock {\em CoRR}, abs/1409.1556, 2014.

\bibitem{sutskever2014sequence}
I.~Sutskever, O.~Vinyals, and Q.~V. Le.
\newblock Sequence to sequence learning with neural networks.
\newblock In {\em Advances in neural information processing systems}, pages
  3104--3112, 2014.

\bibitem{szegedy2014going}
C.~Szegedy, W.~Liu, Y.~Jia, P.~Sermanet, S.~Reed, D.~Anguelov, D.~Erhan,
  V.~Vanhoucke, and A.~Rabinovich.
\newblock Going deeper with convolutions.
\newblock {\em arXiv preprint arXiv:1409.4842}, 2014.

\bibitem{torabi2015using}
A.~Torabi, C.~Pal, H.~Larochelle, and A.~Courville.
\newblock Using descriptive video services to create a large data source for
  video annotation research.
\newblock {\em arXiv preprint arXiv:1503.01070}, 2015.

\bibitem{DBLP:journals/corr/TranBFTP14}
D.~Tran, L.~D. Bourdev, R.~Fergus, L.~Torresani, and M.~Paluri.
\newblock {C3D:} generic features for video analysis.
\newblock {\em CoRR}, abs/1412.0767, 2014.

\bibitem{venugopalan2015sequence}
S.~Venugopalan, M.~Rohrbach, J.~Donahue, R.~Mooney, T.~Darrell, and K.~Saenko.
\newblock Sequence to sequence--video to text.
\newblock {\em arXiv preprint arXiv:1505.00487}, 2015.

\bibitem{DBLP:journals/corr/VenugopalanRDMD15}
S.~Venugopalan, M.~Rohrbach, J.~Donahue, R.~J. Mooney, T.~Darrell, and
  K.~Saenko.
\newblock Sequence to sequence - video to text.
\newblock {\em CoRR}, abs/1505.00487, 2015.

\bibitem{venugopalan2014translating}
S.~Venugopalan, H.~Xu, J.~Donahue, M.~Rohrbach, R.~Mooney, and K.~Saenko.
\newblock Translating videos to natural language using deep recurrent neural
  networks.
\newblock {\em arXiv preprint arXiv:1412.4729}, 2014.

\bibitem{vinyals2014show}
O.~Vinyals, A.~Toshev, S.~Bengio, and D.~Erhan.
\newblock Show and tell: A neural image caption generator.
\newblock {\em arXiv preprint arXiv:1411.4555}, 2014.

\bibitem{xu2015discriminative}
Z.~Xu, Y.~Yang, and A.~G. Hauptmann.
\newblock A discriminative cnn video representation for event detection.
\newblock In {\em Proceedings of the IEEE Conference on Computer Vision and
  Pattern Recognition}, pages 1798--1807, 2015.

\bibitem{yao2015describing}
L.~Yao, A.~Torabi, K.~Cho, N.~Ballas, C.~Pal, H.~Larochelle, and A.~Courville.
\newblock Describing videos by exploiting temporal structure.
\newblock {\em stat}, 1050:25, 2015.

\bibitem{yu2015video}
H.~Yu, J.~Wang, Z.~Huang, Y.~Yang, and W.~Xu.
\newblock Video paragraph captioning using hierarchical recurrent neural
  networks.
\newblock {\em arXiv preprint arXiv:1510.07712}, 2015.

\bibitem{DBLP:journals/corr/ZarembaS14}
W.~Zaremba and I.~Sutskever.
\newblock Learning to execute.
\newblock {\em CoRR}, abs/1410.4615, 2014.

\end{thebibliography}
}

\end{document}